\documentclass[letterpaper]{article} 
\usepackage[preprint]{aaai2027}
\usepackage[hyphens]{url}  
\usepackage{graphicx} 
\usepackage{natbib}  
\usepackage{caption} 
\usepackage{amsmath}
\usepackage{amsfonts}
\usepackage{amssymb}
\usepackage{booktabs}

\title{DSevolve: Enabling Real-Time Adaptive Scheduling on Dynamic Flexible Job Shop with LLM-Evolved Heuristic Portfolios}
\author{
    XinLei Zhou\textsuperscript{\rm 1},
    Jin Huang\textsuperscript{\rm 1},
    Jie Yang\textsuperscript{\rm 1},
    Xinyu Li\textsuperscript{\rm 1}\corresponding,
    Liang Gao\textsuperscript{\rm 1}
}
\affiliations{
    \textsuperscript{\rm 1}School of Mechanical Science and Engineering, Huazhong University of Science and Technology\\
    Wuhan, 430074, PR China\\
    xinyuli@hust.edu.cn
}

\begin{document}

\maketitle

\begin{abstract}
In dynamic flexible job shops, order arrivals, machine breakdowns, and processing-time deviations continually reshape the scheduling state and the priority trade-offs behind dispatching decisions. Dispatching rules are well suited to this setting because they are fast, interpretable, and easy to deploy, and recent LLM-assisted automatic heuristic design further expands their expressiveness by evolving composite priority functions. The key challenge is to make these evolved rule behaviors state-adaptive without losing the rapid response needed for online rescheduling. This paper proposes a dynamic self-evolutionary framework (\textsc{DSevolve}), which separates offline rule-library construction from online state-conditioned rule selection. Offline, an LLM-guided quality-diversity search combines multi-persona seeding, a MAP-Elites behavioral archive, and behavior-guided variation to evolve a library of complementary rules rather than a single elite, and event-level simulation then trains a neural selector to rank the rules by state. Online, after each rescheduling event, a neural selector maps a 22-dimensional state fingerprint to rule scores and dispatches the top-ranked rule within about a second, meeting the response-time requirement after each disruption. Experiments on dynamic instances derived from standard flexible job shop benchmarks show that \textsc{DSevolve} achieves lower mean makespan than individual LLM-evolved rules, classical dispatching rules, and learning-based baselines under a one-active-rule deployment protocol. Trained only on small instances, the selector transfers zero-shot to substantially larger dynamic shops. These results show that state-conditioned selection preserves the speed and interpretability of dispatching rules while improving adaptability.
\end{abstract}


\section{Introduction}
\label{sec:introduction}

The flexible job shop scheduling problem (FJSP) is a classical NP-hard combinatorial optimization problem in manufacturing, where each operation must be assigned to an eligible machine and operations sharing the same machine must be sequenced to minimize the makespan~\cite{brandimarte1993routing}. In real shop floors, however, the scheduling state changes over time. New orders introduce additional operations, machine breakdowns change resource availability, and realized processing times may deviate from planned durations. These events can change the priority trade-offs behind dispatching decisions, turning FJSP into a dynamic problem that requires both low makespan and rapid rescheduling after each event~\cite{ouelhadj2009survey,luo2020dynamic}.

A wide range of methods address different aspects of quality, response speed, and adaptability. Exact formulations based on mixed-integer linear programming or constraint programming can produce high-quality schedules, but repeated re-optimization is often difficult under tight online response requirements. Heuristic dispatching rules (HDRs) remain attractive because they are fast and interpretable~\cite{haupt1989survey}, but a single rule applies the same priority logic across all shop-floor states. Gene expression programming (GEP) and other evolutionary search methods can compose richer reactive rules or policies~\cite{branke2015automated}, yet the learned expression is still fixed once deployed. Deep reinforcement learning (DRL) and neural combinatorial optimization introduce state-dependent decision making, but end-to-end neural schedulers often face scale-transfer and interpretability challenges, while rule-selection agents are bounded by the quality and diversity of their predefined rule sets.

LLM-driven automatic heuristic design (AHD) has recently offered a new route to rule construction. EoH~\cite{liu2024evolution} and ReEvo~\cite{ye2024reevo} show that large language models (LLMs) can synthesize composite priority functions that are more expressive and readable than hand-crafted rules. HSEvo~\cite{dat2025hsevo} and SeEvo~\cite{huang2026automatic} further demonstrate the potential of LLM-assisted evolution for scheduling heuristics. These advances improve the expressiveness of dispatching rules, but their use in dynamic scheduling is still commonly centered on rules evolved offline and then deployed with fixed behavior. Such a rule can perform well on average across training instances, yet different event states within a dynamic trajectory may favor different prioritization behaviors. Re-running LLM-guided evolution after each disruption is also incompatible with the rapid response needed for online rescheduling. The key challenge is therefore to retain the expressiveness of LLM-evolved rules while making their deployment state-adaptive at event time.

To address this challenge, this paper proposes \textsc{DSevolve}, which decouples offline rule-library construction from online state-conditioned rule selection. Offline, an LLM-driven quality-diversity search evolves a behaviorally complementary library of dispatching rules and uses event-level simulations to train a neural selector~\cite{cao2007learning}. Online, at initialization and after each order arrival or machine fault, the selector chooses a rule for the current state and applies it until the next event. On DFJSP instances built from standard benchmarks, \textsc{DSevolve} achieves lower mean makespan than the compared static AHD elite rules, classical dispatching rules, and learning-based baselines. Trained only on small instances, it further transfers zero-shot to considerably larger dynamic shops while retaining its relative advantage.

\paragraph{Contributions}
The main contributions of this paper are summarized as follows:

\begin{itemize}
    \item We build an LLM-assisted quality-diversity rule library for dynamic FJSP, combining multi-persona initialization, a MAP-Elites behavioral archive, and behavior-guided variation operators to retain complementary rules rather than a single elite rule.

    \item We formulate dynamic scheduling as event-level rule selection and train a neural selector from simulation-based labels, using a $22$-dimensional state fingerprint and a ListNet ranking loss to learn the relative ordering of candidate rules rather than a single best-rule label.

    \item We deploy the selector in an event-driven rescheduling loop that selects a rule suited to the current shop-floor state after each event and runs it until the next event, with an optional Top-$K$ look-ahead that enables a tunable trade-off between response time and scheduling quality.
\end{itemize}

\section{Related Work}
\label{sec:related_work}

\subsection{Rules and Learning for Dynamic Scheduling}

DFJSP extends the classical FJSP through interruptions requiring rapid rescheduling, including new order arrivals, machine breakdowns, and processing-time uncertainty~\cite{ouelhadj2009survey,luo2020dynamic}. This response requirement makes repeated optimization or population-based search difficult after each event. HDRs remain practical for speed and interpretability~\cite{haupt1989survey}, but a single rule applies the same priority logic across changing shop-floor states. GEP further automates rule construction by producing strategies for dynamic flexible job shops~\cite{nie2012discover} and feature-selection-based reactive policies~\cite{shady2023feature}. These methods enrich dispatching rules, although the evolved expressions are still fixed once deployed and are usually built from a predefined terminal set.

Recent DRL-based schedulers follow two main directions. End-to-end methods map states directly to scheduling actions~\cite{zhang2020learning,song2022flexible} or iteratively refine complete schedules~\cite{zhang2024deep}, bringing greater flexibility to dynamic scheduling decisions. Their scale transfer and interpretability remain practical concerns. DRL-based rule-selection methods switch among a predefined rule set~\cite{zhao2025deep}. Their effectiveness depends on the quality and diversity of the available rules. In some dynamic settings, carefully designed HDRs can remain competitive with DRL-based methods~\cite{huang2026automatic}, highlighting the continued value of interpretable rule behavior. Other neural approaches learn fast optimization proxies~\cite{kotary2022fast,chen2023two} or target stochastic~\cite{smit2025neural} and structurally uncertain~\cite{zhang2026learning} variants. These approaches broaden the modeling toolbox, but they do not directly address interpretable event-level selection over a rich set of scheduling behaviors. Overall, existing dynamic scheduling methods either rely on a fixed rule behavior or select from a limited predefined action set.

\subsection{LLM-Driven Automatic Heuristic Design}

Using LLMs as evolutionary operators for AHD has recently become a promising paradigm. FunSearch discovers new algorithms via LLM-guided program search~\cite{romera2024mathematical}. AEL couples LLMs with evolutionary search~\cite{liu2023algorithm}. EoH evolves natural-language ideas alongside executable code~\cite{liu2024evolution}, and ReEvo introduces reflective evolution~\cite{ye2024reevo}. HSEvo combines harmony search with genetic operators~\cite{dat2025hsevo}, while NeRM improves search efficiency through nested refinement~\cite{guo2025nested}. In scheduling, this direction is also gaining attention. LSH applies evolved code to static flow-shop, job-shop, and open-shop problems~\cite{yu2026automated}. SeEvo uses a teacher-student mechanism with self-evolving populations for dynamic fuzzy job-shop scheduling~\cite{huang2026automatic}, and LLM4DRD addresses dynamic flexible assembly flow-shop scheduling~\cite{qiu2026llm}.

These studies show that LLM-assisted evolution can produce expressive and readable scheduling heuristics. Much of this line of work focuses on discovering high-performing rules through offline search, often followed by deploying an elite rule or a small set of strong candidates. For dynamic shop floors, this creates a complementary deployment problem. The rule behavior that performs best on average may not be the most suitable one for every event state online, and repeatedly invoking LLM-guided evolution after each disruption is not compatible with rapid rescheduling. Recent variants enrich exploration through MCTS-guided search~\cite{zheng2025monte} and multi-objective evolution~\cite{yao2025multi}. \textsc{DSevolve} complements these efforts. It organizes LLM-evolved heuristics into a quality-diversity rule library and learns an event-level ranking selector that matches the current shop-floor state to a suitable rule behavior, improving state adaptivity while retaining rule interpretability.


\section{Methodology}
\label{sec:method}

\subsection{Problem Formulation}
\label{sec:problem}

We study the DFJSP. In the static FJSP, a set of jobs $\mathcal{J}$ is processed on machines
$\mathcal{M}$. Each job $J_i=(O_{i1},\ldots,O_{in_i})$ is a precedence-ordered
sequence of operations, and each operation $O_{ij}$ is processed on a machine
of its eligible set $\mathcal{M}_{ij}\subseteq\mathcal{M}$, with planned
processing time $\bar p_{ij}^{m}$ on machine $m$. The scheduler decides the machine
assignment of each operation and the processing order on each machine to
minimize the makespan $C_{\max}=\max_{i\in\mathcal{J}} C_{in_i}$, subject to eligibility, precedence ($S_{i,j+1}\ge C_{ij}$), and capacity constraints.

We consider a dynamic shop floor with three sources of uncertainty: new order
arrivals, machine breakdowns, and fuzzy processing times. A new order adds its
operations to the unfinished-task set. A fault makes a machine unavailable, pausing and later resuming its in-progress operation on the same machine.
Fuzzy processing times, modeled with fuzzy scheduling theory~\cite{shi2020multi}, make the realized time $p_{ij}^{m}$ deviate from the
plan by $\delta_{ij}^{m}=p_{ij}^{m}/\bar p_{ij}^{m}-1$. Since these disruptions reshape
dispatching preferences, a single rule is unlikely to remain effective across all states.

We therefore model dynamic scheduling as event-driven rule selection.
Rescheduling is triggered at the event set
\begin{equation}
\mathcal{E}(I)=\{e_0\}\cup\mathcal{E}^{order}(I)\cup\mathcal{E}^{fault}(I),
\label{eq:events}
\end{equation}
including the initial time, the order arrivals, and the fault starts. At event $e$
the shop-floor state is $s_e$. Given a rule library
$\mathcal{R}=\{r_1,\ldots,r_N\}$, we learn a state-to-rule policy
$\pi_\theta:s_e\mapsto r_{k_e}$, where the selected rule is dispatched until the next
event. The objective is
\begin{equation}
\min_{\theta}\;
\mathbb{E}_{I\sim\mathcal{D}}
\big[
C_{\max}\!\left(I;\{\pi_\theta(s_e)\}_{e\in\mathcal{E}(I)}\right)
\big].
\label{eq:objective}
\end{equation}
Thus, the policy selects different rules for different dynamic states instead of
using one fixed rule.

\subsection{Overview}
\label{sec:overview}

\begin{figure*}[t]
\centering
\includegraphics[width=\textwidth]{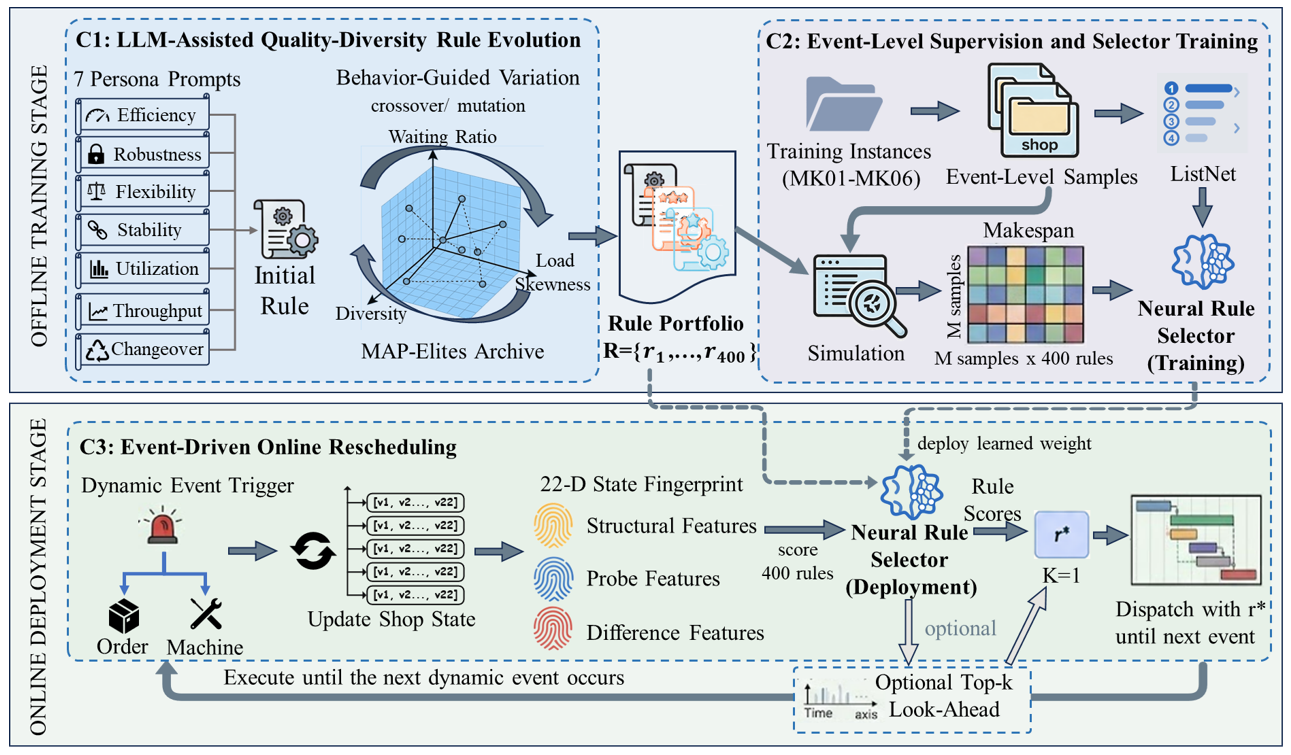}
\caption{Overview of \textsc{DSevolve}.}
\label{fig:pipeline}
\end{figure*}

Figure~\ref{fig:pipeline} gives an overview of \textsc{DSevolve}. The framework operates in three stages: LLM-Assisted Quality-Diversity Rule Evolution, Event-Level Supervision and Selector Training, and Event-Driven Online Rescheduling.

\subsection{LLM-Assisted Quality-Diversity Rule Evolution}
\label{sec:portfolio}

Our first contribution is an LLM-assisted quality-diversity procedure for constructing a rule library.
Instead of converging to one elite rule, we evolve behaviorally complementary
candidates so that the selector can deploy different preferences in different
states.

\paragraph{Rule representation}
Each rule $r_k$ is a composite priority function over a terminal feature vector
$\mathbf{z}(o,m,s)$ combining operation-side and machine-side features. Given the current state $s$, a
candidate operation $o$, and a candidate machine $m$, the rule scores each feasible
operation--machine pair, and the scheduler dispatches the pair with the
smallest score:
\begin{equation}
\begin{aligned}
q_k(o,m\mid s)&=r_k(\mathbf{z}(o,m,s)),\\
(o^\star,m^\star)&=\arg\min_{(o,m)}\; q_k(o,m\mid s),
\end{aligned}
\label{eq:rule}
\end{equation}
where the $\arg\min$ ranges over dispatchable operation--machine pairs
($o\in\mathcal{A}(s)$, $m\in\mathcal{M}_{o}$). A smaller score indicates higher priority.

\paragraph{Multi-persona initialization}
Standard AHD methods initialize from a single prompt. To avoid an early collapse onto a few similar preferences, we prompt the LLM with
seven persona prompts $\mathcal{P}$, each encoding a distinct scheduling
philosophy (e.g., shortest-processing-time first,
bottleneck relief, and look-ahead). The
initial population
\begin{equation}
\mathcal{R}_0=\bigcup_{\rho\in\mathcal{P}}\mathrm{LLM}(\rho).
\label{eq:seed}
\end{equation}
Each persona produces multiple variants with elevated sampling temperature, which gives the search a broad behavioral starting distribution.

\paragraph{Behavior descriptors and MAP-Elites archive}
Retaining rules by average makespan alone drives the population toward a few similar expressions. We therefore associate each rule with a three-dimensional behavioral feature space
\begin{equation}
\mathbf{b}(r_k)=\big[\,b^{load}(r_k),\,b^{wait}(r_k),\,b^{div}(r_k)\,\big],
\label{eq:behavior}
\end{equation}
where $b^{load}(r_k)$ measures workload imbalance across machines, $b^{wait}(r_k)$ captures the average job waiting ratio, and $b^{div}(r_k)$ quantifies the behavioral novelty of the rule relative to the existing archive population~\cite{lehman2011abandoning}. Each rule is mapped to a cell
$c_k=h(\mathbf{b}(r_k))$, where $h$ discretizes the behavior space into a fixed grid of cells, and with fitness $F(r_k)$ equal to the average makespan
over reference instances, the MAP-Elites archive~\cite{mouret2015illuminating}, a representative quality-diversity method~\cite{pugh2016quality}, keeps the best rule per cell:
\begin{equation}
\mathcal{A}[c_k]\leftarrow r_k
\ \ \text{if}\ \ \mathcal{A}[c_k]=\varnothing\ \text{or}\ F(r_k)<F(\mathcal{A}[c_k]).
\label{eq:mapelites}
\end{equation}
This preserves both quality through $F$ and behavioral coverage through cell occupancy.

\paragraph{Behavior-guided variation operators.}
Let the normalized behavior distance between two rules be
$d_{\mathcal{B}}(r_i,r_j)=\|\widetilde{\mathbf{b}}(r_i)-\widetilde{\mathbf{b}}(r_j)\|_2$,
where $\widetilde{\mathbf{b}}$ denotes $\mathbf{b}$ normalized per dimension.
Guided by this distance and by cell crowding, we expand the archive with four
LLM operators that counter premature convergence. \emph{(i)} Complementary
crossover pairs an anchor $r_a$ with the farthest-cell rule
$r_b=\arg\max_{r\in\mathcal{A}} d_{\mathcal{B}}(r_a,r)$ and prompts the LLM to
synthesize their behaviors into a child rule. \emph{(ii)} Self-evolution
refinement compares this child against the better-performing of $r_a$ and
$r_b$ and prompts the LLM to refine it further. \emph{(iii)} Elitist mutation
refines the current global best rule in the archive while preserving its
dominant preference. \emph{(iv)} Contrastive mutation perturbs a rule from a
crowded region toward an opposite tendency, pushing the frontier into sparse
regions. Here, the crowding score of a rule is its average behavior distance to
its $\kappa$ nearest archive neighbors,
\begin{equation}
\mathrm{crowd}(r_k)=\frac{1}{\kappa}\sum_{r\in \mathrm{NN}_\kappa(r_k)} d_{\mathcal{B}}(r_k,r),
\label{eq:crowd}
\end{equation}
so a small $\mathrm{crowd}(r_k)$ marks a densely populated region, and mutation
targets rules with a small $\mathrm{crowd}(r_k)$. The procedure yields a diverse
library of $N=400$ rules, i.e., the final archive $\mathcal{A}$ constitutes the
rule library $\mathcal{R}$ used throughout the rest of the paper.

\subsection{Event-Level Supervision and Selector Training}
\label{sec:selector_training}

Our second contribution constructs supervision from event-level samples and
trains a neural rule selector that scores all rules from a $22$-dimensional state
fingerprint, learning rule rankings instead of a single best label.

\paragraph{Event-level samples}
For each dynamic training instance, we extract samples at the initial time, in a progress state of 30\%, and right after dynamic events. The $n$-th sample has state $s_n$ and type $a_n$
(initial, progress, order, or fault). Randomized dispatch trajectories expose the
selector to diverse shop-floor states rather than a single fixed trajectory.

\paragraph{State fingerprint}
We map a state to a fixed-dimensional vector with three groups,
\begin{equation}
\mathbf{x}_n=g(s_n)=\big[\,\mathbf{x}^{struct}_n;\;\mathbf{x}^{probe}_n;\;\mathbf{x}^{diff}_n\,\big]\in\mathbb{R}^{22},
\label{eq:fingerprint}
\end{equation}
where $\mathbf{x}^{struct}_n$ are structural features of the current state,
$\mathbf{x}^{probe}_n$ are lightweight look-ahead features obtained by quickly
rolling out a simple heuristic, and $\mathbf{x}^{diff}_n$ are dual-probe
difference features. Concretely, rolling out an SPT and an LPT probe to completion
yields two estimated makespans $\widehat C_{\max}^{SPT}(s_n)$ and
$\widehat C_{\max}^{LPT}(s_n)$, whose normalized gap
\begin{equation}
\Delta^{probe}(s_n)=\frac{\widehat C_{\max}^{LPT}(s_n)-\widehat C_{\max}^{SPT}(s_n)}{\widehat C_{\max}^{SPT}(s_n)+\epsilon}
\label{eq:probe}
\end{equation}
measures how sensitive the state is to the rule choice. A large gap signals that
the rule choice matters, and a small gap that it does not.

\paragraph{Simulation-based labels}
For each sample, we simulate every rule from the current state to completion,
which we refer to as a rule rollout, and obtain
\begin{equation}
\begin{aligned}
\mathbf{y}_n&=(y_{n,1},\ldots,y_{n,N}),\\
y_{n,k}&=C_{\max}\!\left(\operatorname{Rollout}(s_n,r_k)\right),
\end{aligned}
\label{eq:labels}
\end{equation}
and a training set $\mathcal{D}_{train}=\{(\mathbf{x}_n,\mathbf{y}_n,a_n)\}_{n=1}^{M}$.
Compared with a hard label that keeps only the best rule, the full makespan vector
preserves the complete ordering and expresses the gaps among near-optimal rules,
which yields a more stable learning signal. Crucially, in the offline simulation used for label construction, each rollout is evaluated under the realized processing-time scenario associated with the instance, so $\mathbf{y}_n$ reflects the
makespan actually attained under fuzzy processing times. Thus, although the
selector receives no explicit deviation feature, its target already encodes
robustness to the plan-to-actual gap.

\paragraph{Event-conditioned MLP and ListNet loss}
The selector outputs scores
$\mathbf{s}_n=f_\theta([\mathbf{x}_n;\mathbf{e}(a_n)])$, which induce the policy
$\pi_\theta$ from Eq.~\ref{eq:objective} by selecting the lowest-scoring rule at
each event, where $\mathbf{e}(a_n)$ is
a learnable event-type embedding that conditions the network on the event type
and a smaller score means a better rule. Casting selection as learning-to-rank, we convert makespans and scores into top-one probability distributions over the valid rule set $\mathcal{V}_n$ (with failed rules masked and $T$ denoting a temperature):
\begin{equation}
\begin{aligned}
p_{n,k}&=\frac{\exp(-y_{n,k}/T)}{\sum_{j\in\mathcal{V}_n}\exp(-y_{n,j}/T)},\\[3pt]
\hat p_{n,k}&=\frac{\exp(-s_{n,k}/T)}{\sum_{j\in\mathcal{V}_n}\exp(-s_{n,j}/T)}.
\end{aligned}
\label{eq:dist}
\end{equation}
We then minimize a weighted ListNet loss,
\begin{equation}
\mathcal{L}=-\sum_{n}w(a_n)\sum_{k\in\mathcal{V}_n}p_{n,k}\log\hat p_{n,k},
\label{eq:listnet}
\end{equation}
where $w(a_n)$ weights the order and fault samples. A single forward pass selects
$k^\star=\arg\min_k s_k$ with millisecond-level neural inference. Because the loss is fit to the full ranking rather than a single best-rule label, the selector effectively learns to recognize the whole cluster of near-optimal rules for a given state, not just the identity of the single top one. Consequently, even when $k^\star$ misses the true optimum among the $N=400$ candidates, it still lands within this near-optimal cluster and remains well matched to the current state, keeping the risk of a poorly adapted selection low.

\subsection{Event-Driven Online Rescheduling}
\label{sec:online}

Our third contribution deploys the trained selector in an event-driven loop. The
main method selects one rule per event without additional simulation. A Top-$K$
look-ahead is an optional enhancement.

\paragraph{Stage-1 direct selection}
At each event $e$, the system extracts $\mathbf{x}_e=g(s_e)$, selects
$k_e=\arg\min_k f_\theta(\mathbf{x}_e,a_e)_k$, and dispatches $r_{k_e}$ until the
next event:
\begin{equation}
s_e \xrightarrow{\,g\,} \mathbf{x}_e \xrightarrow{\,f_\theta\,} r_{k_e}
\xrightarrow{\,\text{dispatch}\,} s_{e+1}.
\label{eq:online}
\end{equation}
We denote this default mode \textsc{DSevolve}.
Rescheduling changes only not-yet-started operations and never interrupts an
in-progress one.

\paragraph{Optional Top-$K$ look-ahead}
Alternatively, the selector keeps its top $K$ candidates and validates each one with a short forward rollout of the schedule:
\begin{equation}
\begin{aligned}
\mathcal{K}_e&=\operatorname{TopK}(-\mathbf{s}_e,K),\\
k_e^{TopK}&=\arg\min_{k\in\mathcal{K}_e}\widehat C_{\max}(s_e,r_k).
\end{aligned}
\label{eq:topk}
\end{equation}
where $\widehat C_{\max}(s_e,r_k)$ is the makespan estimated from a short
rollout of rule $r_k$ from state $s_e$. This improves quality at higher online cost, so we treat $K$ as a
parameter-sensitivity study rather than part of the main method.

\section{Experiments}
\label{sec:experiments}

We evaluate the proposed method in terms of scheduling quality, generalization, component contributions, Top-$K$ sensitivity, and online response cost.

\subsection{Experimental Setup}
\label{sec:exp_setup}

\paragraph{Dataset}
We construct DFJSP instances from the standard Brandimarte benchmarks with new order arrivals, machine breakdowns, and fuzzy processing times. Specifically, MK01--MK06 are used to build 500 small-scale training instances, split into 425 training and 75 validation instances. MK07--MK10 are used to build 320 medium-scale test instances, and MK11--MK15 are used to build 400 large-scale generalization instances. The base job count increases from 10 in training to 20 in testing and 30 in generalization. After order insertions, the largest generalization instances contain up to 330 jobs. This setting evaluates zero-shot transfer from small training instances to medium-scale and large-scale dynamic instances.

\paragraph{Model}
We use Qwen-Plus to generate a library of 400 composite dispatching rules. All baselines that require LLMs also use Qwen-Plus to avoid confounding model differences. The rule-selection network maps a 22-dimensional state fingerprint to a 400-dimensional score vector. The main method, denoted as \textsc{DSevolve}, directly selects the top-ranked rule after each rescheduling event.

\paragraph{Baselines}
We compare with three groups of baselines: (1) AHD frameworks, including EoH~\cite{liu2024evolution}, ReEvo~\cite{ye2024reevo}, and HSEvo~\cite{dat2025hsevo}. Each AHD framework evolves its own globally best elite rule on the training instances and then uses this rule throughout testing. (2) Seven common dispatching heuristics: FIFO, SPT, LPT, MOR, LOR, MWKR, and LWKR. (3) Evolutionary and learning-based methods: DRL-GP~\cite{zhao2025deep}, GEP-Nie~\cite{nie2012discover}, and GEP-Shady~\cite{shady2023feature}.

\paragraph{Metrics}
The primary quality metrics are the mean and median final makespan. We also report absolute and relative gaps. For instance $i$, let $C_i^m$ be the makespan of method $m$, and let $C_i^\star=\min_m C_i^m$ be the best makespan among the methods compared in the corresponding table. We define
\begin{equation}
\operatorname{AbsGap}(m)=\frac{1}{|\mathcal{I}|}\sum_{i\in\mathcal{I}}(C_i^m-C_i^\star),
\end{equation}
\begin{equation}
\operatorname{RelGap}(m)=\frac{1}{|\mathcal{I}|}\sum_{i\in\mathcal{I}}\frac{C_i^m-C_i^\star}{C_i^\star}\times 100\%.
\end{equation}
We also report the win rate of \textsc{DSevolve} against each baseline, defined as the fraction of instances on which it achieves a strictly lower makespan.

\subsection{Main Results}
\label{sec:exp_main}

\begin{table*}[t]
\centering
\begin{tabular}{lrrrrrl}
\toprule[1.2pt]
Method & Mean & Median & AbsGap & RelGap(\%) & Win(\%) & W/T/L \\
\midrule
EoH & 5422.40 & 2837.52 & 258.44 & 6.10 & 72.50 & 232/0/88 \\
HSEvo & 5482.98 & 2821.29 & 319.02 & 6.43 & 71.25 & 228/0/92 \\
ReEvo & 5493.66 & 2908.61 & 329.69 & 7.21 & 71.88 & 230/0/90 \\
\midrule
FIFO & 8658.53 & 4515.22 & 3494.57 & 69.67 & 98.12 & 314/0/6 \\
SPT & 7539.70 & 4035.37 & 2375.74 & 51.63 & 98.75 & 316/0/4 \\
LPT & 7573.48 & 4078.17 & 2409.51 & 52.28 & 98.44 & 315/0/5 \\
MOR & 7749.60 & 4134.02 & 2585.64 & 52.56 & 98.12 & 314/0/6 \\
LOR & 7488.78 & 4068.14 & 2324.82 & 50.58 & 97.19 & 311/0/9 \\
MWKR & 7706.97 & 4055.04 & 2543.01 & 52.82 & 97.81 & 313/0/7 \\
LWKR & 7493.53 & 4033.67 & 2329.57 & 50.76 & 97.19 & 311/0/9 \\
\midrule
DRL-GP & 5443.55 & 2908.18 & 279.59 & 7.93 & 75.62 & 242/0/78 \\
GEP-Nie & 5938.82 & 3222.86 & 774.86 & 17.40 & 89.69 & 287/0/33 \\
GEP-Shady & 5487.77 & 2928.55 & 323.81 & 7.38 & 72.81 & 233/0/87 \\
\midrule
\textbf{DSevolve} & \textbf{5296.75} & \textbf{2811.12} & \textbf{132.79} & \textbf{2.74} & - & - \\
\bottomrule[1.2pt]
\end{tabular}
\caption{Main comparison. AbsGap and RelGap are computed against the best makespan among all methods in this table. Win and W/T/L denote the win rate and win/tie/loss counts of \textsc{DSevolve} against each method. Best results are in bold.}
\label{tab:main_comparison}
\end{table*}

Table~\ref{tab:main_comparison} shows that \textsc{DSevolve} achieves the lowest mean makespan, 5296.75, with a RelGap of 2.74\%. Against static AHD elite
rules, it wins on 72.50\%, 71.25\%, and 71.88\% of instances over EoH, HSEvo, and
ReEvo, respectively, showing the limitation of one fixed elite rule in dynamic
shops.

The gap to each baseline also reveals a consistent hierarchy. All seven
classical dispatching heuristics have a RelGap above 50\%, while every
composite-rule baseline, whether evolved by an LLM (EoH, HSEvo, ReEvo), GEP (GEP-Nie, GEP-Shady), or DRL-GP, remains
within 20\%. This split suggests that combining multiple scheduling features
into one rule, rather than ranking jobs by a single attribute, already closes
most of the gap to a well-tuned dispatching policy. The remaining gap between
\textsc{DSevolve} and these composite-rule baselines therefore isolates a
different source of improvement, coming not only from rule complexity,  but also
the ability to switch among complementary rules as the shop-floor state
changes.

\textsc{DSevolve} wins over all classical dispatching rules in more than
97\% of test instances. Compared to learning-based baselines, our
method also maintains a clear advantage, with win rates of 75.62\%, 89.69\%,
and 72.81\% against DRL-GP, GEP-Nie, and GEP-Shady, respectively. These
results show that event-level deep rule selection outperforms static elite
rules, classical heuristics, and learning-based baselines under a
fair single-rule protocol.

\subsection{Generalization}
\label{sec:exp_generalization}

\begin{table}[t]
\centering
\begin{tabular}{lrrr}
\toprule[1.2pt]
Method & Mean & Median & RelGap(\%) \\
\midrule
EoH & 7250.09 & \textbf{4569.78} & 5.52 \\
HSEvo & 7383.21 & 4736.70 & 6.23 \\
ReEvo & 7367.61 & 4928.55 & 6.20 \\
\midrule
FIFO & 12428.30 & 7404.08 & 79.26 \\
SPT & 10894.69 & 7073.06 & 61.39 \\
LPT & 11013.68 & 7078.64 & 63.33 \\
MOR & 11229.44 & 7048.68 & 63.70 \\
LOR & 10888.19 & 7045.81 & 61.59 \\
MWKR & 11212.88 & 7277.54 & 63.94 \\
LWKR & 10874.14 & 7159.87 & 61.23 \\
\midrule
\textbf{DSevolve} & \textbf{7113.22} & 4683.30 & \textbf{2.19} \\
\bottomrule[1.2pt]
\end{tabular}
\caption{Generalization results. The model is not retrained on the generalization set. Best results are in bold.}
\label{tab:generalization}
\end{table}

Because the selector is trained only on the small MK01--MK06 instances, we evaluate zero-shot transfer at two scales: the MK07--MK10 test set, used for the full comparison, ablation, Top-$K$, and runtime analyses, and the larger MK11--MK15 generalization set, which reaches up to 330 jobs after order insertions and tests whether the advantage persists as the scale gap widens. Table~\ref{tab:generalization} shows that without retraining, \textsc{DSevolve} still achieves the best mean makespan and the lowest RelGap, winning against EoH, HSEvo, and ReEvo on 75.00\%, 75.50\%, and 77.75\% of instances and beating every heuristic dispatching rule on all 400 generalization instances. 

Comparing RelGap across the two scales reveals a divergent trend that mirrors the composite-versus-single-feature split in Table~\ref{tab:main_comparison}. Every classical heuristic's RelGap grows at the larger scale, for example from 69.67\% to 79.26\% for FIFO, while the RelGap of every composite-rule baseline and of \textsc{DSevolve} shrinks instead, for example from 6.10\% to 5.52\% for EoH and from 2.74\% to 2.19\% for \textsc{DSevolve}. A rule based on a single feature likely compounds more errors as the number of operations grows, whereas composite rules and state-conditioned selection remain more robust to scale, consistent with a state fingerprint built from ratios and normalized differences rather than raw counts (Eq.~\ref{eq:fingerprint}). \textsc{DSevolve} trails EoH slightly on the median (4683.30 vs. 4569.78) yet leads on the mean (7113.22 vs. 7250.09), and since the mean is pulled more by the largest instances, this pattern suggests the advantage over EoH concentrates on the harder, larger cases where state-conditioned selection should matter most. The learned selector transfers from small training instances to larger dynamic shops while retaining its advantage over static AHD elites and HDRs.

\begin{table}[t]
\centering
\begin{tabular}{lrrr}
\toprule[1.2pt]
Method & Mean & Median & Win(\%) \\
\midrule
\textbf{DSevolve} & \textbf{5296.75} & \textbf{2811.12} & - \\
DSevolve-Single & 5396.52 & 2879.56 & 72.19 \\
w/o Persona & 5519.84 & 2818.00 & 71.25 \\
w/o Feature Space & 5489.75 & 3084.13 & 77.81 \\
\bottomrule[1.2pt]
\end{tabular}
\caption{Ablation results. Win denotes the instance-level win rate of the full \textsc{DSevolve} against each variant. Best results are in bold.}
\label{tab:ablation}
\end{table}

\subsection{Ablation Study}
\label{sec:exp_ablation}

\begin{table*}[!t]
\centering
\begin{tabular}{rrrrrrrrr}
\toprule[1.2pt]
K & Mean & Median & WinK1(\%) & Decision & Feature & Selector & SerialVal & ParallelVal \\
\midrule
1 & 5296.75 & 2811.12 & - & \textbf{1165.99} & 1164.77 & 0.88 & \textbf{0.00} & \textbf{0.00} \\
5 & 5101.83 & 2773.70 & 71.56 & 1826.52 & 1057.85 & \textbf{0.81} & 3837.29 & 767.46 \\
10 & 5068.71 & 2773.80 & \textbf{74.06} & 2036.90 & 1112.79 & 0.91 & 9227.55 & 922.75 \\
20 & 5042.95 & \textbf{2719.53} & \textbf{74.06} & 1986.58 & 1064.99 & 0.84 & 18406.43 & 920.32 \\
30 & \textbf{5036.99} & \textbf{2719.53} & 73.75 & 2173.18 & \textbf{1047.28} & 0.84 & 33739.05 & 1124.63 \\
\bottomrule[1.2pt]
\end{tabular}
\caption{Top-$K$ quality--time trade-off. All time columns are in milliseconds. WinK1 is the win rate against $K=1$, and Decision uses parallel validation latency. Best results are in bold.}
\label{tab:topk_runtime}
\end{table*}

Table~\ref{tab:ablation} separates the contributions of rule library construction and online rule selection. \textsc{DSevolve}-Single mirrors the deployment protocol of the AHD baselines, drawing on the same diverse rule library but collapses to a single rule chosen by training performance, applying it throughout each instance. Its mean makespan is 5396.52, compared with 5296.75 for the full method, and \textsc{DSevolve} wins on 72.19\% of the instances. This shows that the improvement comes not only from containing a strong rule in the library but also from event-level, state-dependent rule selection.

Removing multi-persona initialization increases the mean makespan to 5519.84, while removing the behavior feature space increases it to 5489.75, with the full method winning against these two variants on 71.25\% and 77.81\% of instances, respectively. Because the MK01--MK06 skeletons expand into instances of very different sizes, the median mainly reflects typical, smaller instances while the mean is pulled more by the largest ones. w/o Persona keeps a median close to the full method (2818.00 vs. 2811.12) but a much higher mean, so its degradation concentrates on larger instances, whereas w/o Feature Space shows a comparably sized increase in both statistics, pointing to a more uniform loss across scales. Multi-persona seeding therefore matters most on larger, harder instances, while the behavior-guided archive contributes more evenly.

\subsection{Top-$K$ Sensitivity and Online Response Cost}
\label{sec:exp_topk_runtime}

Top-$K$ look-ahead is an optional enhancement. The network first keeps the top $K$ rules, then simulates these candidates from the current state and chooses the one with the smallest rollout makespan. This validation assumes the selected rule runs until the end of the instance, whereas online execution may later trigger reselection after order arrivals or machine faults, so Top-$K$ is not guaranteed to beat $K=1$ on every instance even though it improves average quality. Table~\ref{tab:topk_runtime} shows the mean makespan decreasing from 5296.75 at $K=1$ to 5042.95 at $K=20$ and 5036.99 at $K=30$, with Top-$K$ winning against $K=1$ on about 72\%--74\% of instances. Gains shrink quickly as $K$ grows, from 194.92 in mean makespan between $K=1$ and $K=5$ to under 6 between $K=20$ and $K=30$, since deeper-ranked candidates are already scored lower by the trained selector and rarely turn out best once rolled out.

All online simulations run on CPUs. For $K=1$, \textsc{DSevolve} makes 7.66 rule decisions per instance on average, averaging 1165.99 ms per decision. The neural forward pass takes only 0.88 ms. The main cost comes from state-feature construction and probe evaluation, which together take 1164.77 ms on average. Although the current bottleneck is state representation rather than neural inference, the total response time remains around one second, which meets the timing requirement of online rescheduling after order arrivals or machine faults.

For Top-$K$, the extra cost mainly comes from candidate validation. The Decision column in Table~\ref{tab:topk_runtime} uses the parallel validation time, and evaluating candidates serially instead would increase the response time accordingly. In deployment, candidates can be validated in parallel, with the response time approximated by dividing the serial validation budget by the number of available workers. With 20 parallel workers, $K=20$ takes 1986.58 ms per decision on average, including 920.32 ms for parallel validation, against a serial validation budget of 18406.43 ms. Even with Top-$K$, the wall-clock response remains around two seconds per rescheduling decision, still practical for dynamic shop-floor production, though the default \textsc{DSevolve} mode or a smaller $K$ is preferable when only serial execution is available. The framework therefore offers a tunable quality-time trade-off, with the default mode supporting fast online response and Top-$K$ suited to settings with more computation and a stronger preference for scheduling quality.

\section{Conclusion}
\label{sec:conclusion}

This paper studied DFJSP from the perspective of state-adaptive rule deployment. While LLM-driven AHD has made dispatching rules more expressive, dynamic shop floors may require different rule behaviors as order arrivals, machine breakdowns, and processing-time deviations reshape the scheduling state. We introduced \textsc{DSevolve}, which builds a behaviorally complementary rule library offline through LLM-guided quality-diversity search and trains a neural selector to choose a suitable rule after each rescheduling event. The selected rule is then applied until the next event, preserving the fast response of dispatching-based scheduling. Across DFJSP instances built from standard benchmarks, \textsc{DSevolve} achieved lower mean makespan than the compared static AHD elite rules, classical dispatching rules, and learning-based baselines. The selector also transferred zero-shot from small training instances to substantially larger dynamic shops while retaining its relative advantage. Since the online stage only requires state-feature extraction and a lightweight selector evaluation rather than repeated evolutionary search, \textsc{DSevolve} completes each rescheduling decision within about a second, matching the response-time requirement.

\bibliography{refs}


\end{document}